\UseRawInputEncoding
\documentclass[aoas,preprint]{imsart}

\RequirePackage[OT1]{fontenc}
\RequirePackage{amsthm,amsmath}
\usepackage{amsfonts}
\RequirePackage[authoryear]{natbib}
\RequirePackage[colorlinks,citecolor=purple,urlcolor=purple]{hyperref}
\usepackage{graphicx}
\usepackage{xcolor}
\usepackage{bbm}
\usepackage{algorithm}
\usepackage[noend]{algpseudocode}
\usepackage{caption}

\definecolor{green}{rgb}{0.0, 0.5, 0.0}

\makeatletter
\def\BState{\State\hskip-\ALG@thistlm}
\makeatother

\newcommand{\by}{{\mathbf{y}}}

\newcommand{\bP}{{\mathbf{P}}}

\newcommand{\bX}{{\mathbf{X}}}

\newcommand{\balpha}{\boldsymbol{\alpha}}

\newcommand{\bbeta}{\boldsymbol{\beta}}

\newcommand{\bmu}{\boldsymbol{\mu}}

\startlocaldefs
\numberwithin{equation}{section}
\theoremstyle{plain}

\endlocaldefs

\begin{document}

\begin{frontmatter}
\title{Bayesian Learning of Play Styles in Multiplayer Video Games}
\runtitle{Learning of Video Game Play Styles}

\begin{aug}
\author{\fnms{Aline} \snm{Normoyle}} \and
\author{\fnms{Shane T.} \snm{Jensen}}

\affiliation{Bryn Mawr College and University of Pennsylvania}

\address{Aline Normoyle\\
Department of Computer Science \\ 
Bryn Mawr College \\
101 North Merion Ave\\
Bryn Mawr, PA 19010\\
Email: \href{mailto:anormoyle@brynmawr.edu}{anormoyle@brynmawr.edu}}

\address{Shane T. Jensen\\
Department of Statistics \\
The Wharton School \\ 
University of Pennsylvania \\
Academic Research Building, Room 415 \\
265 South 37th Street\\
Philadelphia, PA 19104\\
Email: \href{mailto:stjensen@wharton.upenn.edu}{stjensen@wharton.upenn.edu}}

\end{aug}

\begin{abstract}
The complexity of game play in online multiplayer games has generated strong interest in modeling the different {\it play styles} or strategies used by players for success. 
We develop a hierarchical Bayesian regression approach for the online multiplayer game Battlefield 3 where performance is modeled as a function of the roles, game type, and map taken on by that player in each of their matches.   We use a Dirichlet process prior that enables the clustering of players that have similar player-specific coefficients in our regression model, which allows us to discover common {\it play styles} amongst our sample of Battlefield 3 players.  This Bayesian semi-parametric clustering approach has several advantages: the number of common play styles do not need to be specified, players can move between multiple clusters, and the resulting groupings often have a straight-forward interpretations.  We examine the most common play styles among Battlefield 3 players in detail and find groups of players that exhibit overall high performance, as well as groupings of players that perform particularly well in specific game types, maps and roles.  We are also able to differentiate betweeen players that are stable members of a particular play style from hybrid players that exhibit multiple play styles across their matches.   Modeling this landscape of different play styles will aid game developers in developing specialized tutorials for new participants as well as improving the construction of complementary teams in their online matching queues. 
\end{abstract}


\end{frontmatter}

\section{Introduction}

Online multiplayer games are a tremendously popular part of the video game industry which now has revenues in excess of other entertainment industries such as film and sports \citep{Wit20}.   These games typically pit either individual players or teams of players against each other in an adversarial game scenario (player versus player or PVP for short) with particular goals, such as capturing a flag or a match to the death.  

In particular, we will focus on the online game Battlefield 3 produced by Electronic Arts.   In Battlefield 3, online players are segregated into two teams by a queuing algorithm and then both teams are placed on a military battlefield with a designated goal determined by the game type, such as capture the flag or deathmatch.  Players must coordinate with their teammates towards their goal while being opposed by players on the other team.    Individual player success is measured by several outcomes, including number of kills, number of deaths and a game score that is influenced by the type of game.  

In games such as Battlefield 3, the complexity of the battle scenarios allows for many different paths to victory and so 
a primary interest is isolating common play styles and strategies employed by participants.  Identifying subgroups of the player population that employ similar play styles can help game designers to tailor tutorials for newer players in a style-specific way as well as aid queuing procedures to construct teams of players with complementary styles.   

From a statistical perspective, the clustering of players into groups with similar play styles can serve the additional role of dimension reduction which is an important factor given the size of most game play datasets.   As we will see in Section~\ref{data}, our Battlefield 3 application consists of the results from half a million PVP matches involving over a thousand players that have been designated by the company as ``focal" players.  The amount of data available for all players of Battlefield 3 or similar games is much larger.  

Clustering has been recognized previously as an important tool for understanding player preferences and their interactions with the game~\citep{GameAnalytics2013}.  \cite{Drachen12GunsSwords} used clustering to identify player preferences for using vehicles over direct combat in {\it Tera} and {\it Battlefield 2: Bad Company 2}.  \cite{Drachen09SOMTombRaider}, \cite{Sifa13TombRaiderArchtype} and \cite{Sifa13TombRaiderArchtype} modeled styles of game play and how they evolve in  {\it Tomb Raider: Underworld}.   

\cite{Thurau11WOWLevels} clustered players by how their character level changed over time and \cite{Thurau10Guilds} employed clustering to understand how guilds evolve over time in {\it World of Warcraft}.  \cite{Holmgard13SuperMario} used a hierarchical clustering method to group players based on how they differed from a ``perfect'' automated player in {\it Super Mario Brothers}.  \cite{Nogueira14FuzzyAffective} used clustering for modeling how player emotions related to game events in the survival horror game {\it Vanish}.  \cite{Bauckhage14BeyongHeatmaps} identified hotspots of player activities based on clusters of player trajectories in {\it Quake III}.  \cite{Tychsen08Personas} defined design-based clusterings of players called personas in {\it Hitman: Blood Money}.

The most common approach among these previous studies is the clustering of players directly on outcomes of matches, such as kills, deaths or the match score.   In contrast, we are more interested in discovering different {\it player styles} rather than just differentiating players based on their overall ability.    We are defining {\it play styles} as the choices that players make in terms of the roles, game type and map type that they prefer, and how they perform under their choices.    Thus we will focus on clustering players based on how their choices affect the Battlefield 3 game outcome, rather than clustering on game outcomes directly.   

Our approach to play styles is based upon a regression model with game score as a function of the roles, game type, and map chosen by each player in each match.  We also include the rank of each player in this regression model to account for differences in player ability.   We estimate both global and player-specific coefficients on each of these covariates.  The player-specific coefficients can also be interpreted as measures of how well that player performs relative to the global performance across all players under specific role/game/map choices. 

The set of player-specific coefficients is what defines a player's {\it style}: how each role/game/map choice by the player relates to their team's performance in the match.   We then discover common play styles across players by employing a semi-parametric Bayesian clustering approach based on a Dirichlet process, which allows us to discover groups of players that have similar coefficients.   The Dirichlet process \citep{Ferguson74PriorDistributions}, as reviewed in \cite{Muller04NonparametricBayesian}, is the basis for many model-based clustering approaches~\citep{Griffin06OrderBased,Teh06DP}.  


Relative to k-means, our choice of a model-based Bayesian clustering procedure has the important advantage that the number of clusters, i.e. unique {\it player styles}, does not have to be pre-specified.   We employ k-means clustering to provide a good initialization for our model estimation, but the number of clusters can grow (or shrink) as the algorithm proceeds based on whether extra (or fewer) clusters are needed to provide the best explanation of the observed data.

Our semi-parametric Bayesian clustering approach is also related to the DP-means procedure of \cite{Kulis12DPMeans}.  However, our focus is on clustering player-specific regression coefficients (that define latent {\it player styles}) rather than Gaussian means.   More importantly, our Markov Chain Monte Carlo implementation does not produce a ``hard clustering" that we would get from DP-means.   Rather, players are able to switch between clusters (play styles) in our approach.  

Our ``soft clustering" strategy that allows players to move between clusters accommodates two types of potential variability exhibited by players in Battlefield 3.  First, certain players may have a play style that is a true hybrid between two (or more) common play styles shared by many players. Second, certain players can transition to new play styles over time as they continue to play the game and so they should be allowed to move between play style clusters as we observe more matches for that player.

We describe the data from our Battlefield 3 application in Section~\ref{data}.   Our regression model for game play data and semi-parametric Bayesian clustering model for play styles is outlined in Section~\ref{model}.    

In Section~\ref{application}, we apply our model to the large Battlefield 3 dataset described in Section~\ref{data}.  We first analyze a subset of our data with the static version of our model and then apply the adaptive version of our model to the larger complete dataset of Battlefield 3 matches.   We conclude with a discussion in Section~\ref{discussion}.  

\section{Battlefield 3 Data}  \label{data}

Data from Battlefield 3 was provided by Electronic Arts through the Wharton Customer Analytics Initiative.  Battlefield 3 is first-person military-themed online game which allows players to use a variety of weapons and vehicles in diverse environments across the globe, ranging from tight urban landscapes to open desert.   Our data consists of 515,605 player-versus-player match logs taken from 1221 players. 

In each match, online players are segregated into teams that are placed on a chosen map with a particular goal determined by the game type.   There are 9 possible roles (e.g. assault, support, recon, engineer along with the vehicle roles of armored land, unarmored land, helicopter, boat, and jet), 17 possible game types (e.g. conquest, rush, and death match) and 30 possible maps (e.g Grand Bazaar, Noshahr Canals, or Operation Metro).   For each match, our data contains information about each player's chosen roles,  map, and game type as well as each player's rank (a measure of their progression).  

Each player may choose more than one role in a match and they might join a match late or quit a match early.  We only consider player matches for which the player played more than 5 minutes and accumulated more than 100 points.  Shorter matches with less than 100 points tend to be the matches where a player quit early or joined late and thus did not have much time to exhibit their play style in the game.  

There are several different match outcomes that we can consider: total match score, score just from combat, number of kills and number of deaths.  In our analysis, we focus on total score as an outcome variable since it includes points due to both combat and objectives and there is a better indicator of how players performed overall in the match regardless of whether their team won or lost.

\section{Model and Estimation} \label{model}

Our approach begins with a linear regression model on each player's total score as a function of the character rank, roles, game type, and map. The set of player-specific coefficients from this regression defines a player's {\it play style}: how each role and game and map choice relates to their performance.   We then employ a Bayesian nonparametric clustering to find subsets of players that all share similar play styles.   

\subsection{Regression Model for Game Play Data} \label{regression}

The first component of our method is a regression model of the total score for player $j$ in each match $m$ as a function of that player's rank as well as their chosen roles.   

As mentioned in Section~\ref{data}, we focus on the total score, $\text{Score}_{jm}$, as the outcome for player $j$ in match $m$ since it combines performance from both combat and objectives, so it should be a more comprehensive measure of performance than player kills or deaths.   We log transform $\text{Score}_{jm}$ so that our residual errors $\varepsilon_{jm}$ more closely match the assumption of being normally distributed.

Let us first consider the following linear model, 
\begin{align}\label{globalmodel}
\log \text{Score}_{jm} \,\, = \,\, y_{jm} \,\, = \,\, \alpha_{0} \, + \, &\alpha_{1} \, \text{Rank}_{jm} \, + \, \balpha_{2} \, \text{Roles}_{jm} \, + \nonumber \\
                              &\balpha_{3} \, \text{Games}_{jm} \, + \, \balpha_{4} \, \text{Maps}_{jm} \, + \, \varepsilon_{jm}
\end{align}
where $\varepsilon_{jm} \, \sim \, {\rm Normal} (0 \, , \, \sigma^2)$.  Each player's $\text{Rank}_{jm}$ is a non-negative integer value, ranging from 0 to 145 that measures the player's progression in previous games up to that point.  We include this covariate since players with higher ranks typically achieve higher scores due to their greater experience with game play.    

The other covariates $\text{Roles}_{jm}$, $\text{Games}_{jm}$ and $\text{Maps}_{jm}$ are indicator variables for the roles, game type, and map that are used by player $j$ in match $m$.   For example, if player $j$ used a helicopter in match $m$, the corresponding role indicator variable would be 1; otherwise, it is zero.   The coefficients on these indicators  represent how players perform on average in particular roles/games/maps beyond what would be expected just based on their player rank. 

Note that the coefficients $\balpha$ in (\ref{globalmodel}) are not indexed by player $j$ and so represent the global effects of each covariate estimated over all players in the data.  These global coefficients provide insight into the relative importance of rank, roles, game types and maps on the performance of all players.  However, model (\ref{globalmodel}) is insufficient for our goal of estimating differences between players in terms of play styles.  

We build player heterogeneity into our model by adding player-specific coefficients, 
\begin{align}\label{playermodel}
 y_{jm} \,\, = \,\,  &\alpha_{0} \, + \, \beta_{j0} \, + \, \alpha_{1} \, \text{Rank}_{jm} \, +  \, \beta_{j1} \, \text{Rank}_{jm} \, + \nonumber \\ 
 & \balpha_{2} \, \text{Roles}_{jm} \, + \, \balpha_{3} \, \text{Games}_{jm} \, + \, \balpha_{4} \, \text{Maps}_{jm} \, +  \\
  &  \bbeta_{j2} \, \text{Roles}_{jm} \, + \, \bbeta_{j3} \, \text{Games}_{jm} \, + \, \bbeta_{j4} \, \text{Maps}_{jm} \, + \, \varepsilon_{jm} \nonumber 
\end{align}
The coefficients $\alpha_{0}$ and  $\alpha_{1}$ capture the average performance across all players and how, on average, player performance increases as their Battlefield 3 rank increases.   
The player-specific coefficients $\beta_{j0}$ and  $\beta_{j1}$ capture how the overall performance of player $j$ differs from the average player and how player $j$ differs from average in their change in performance as their Battlefield 3 rank increases.   The remaining player-specific coefficients $(\beta_{j2}, \beta_{j3}, \ldots)$ represent how the performance of player $j$ differs from the average performance of players in the specific role, map, and game types encountered in each of their matches.  We denote with the vector $\bbeta_j$ all the player-specific coefficients for player $j$.

A player's performance is also affected by the play of their opponents, and so ideally we would incorporate opponent choices into our model.  Unfortunately our available data only consists of the choices/performance of our set of 1221 focal players, not their match-specific opponents.  

\subsection{Dirichlet Process Prior for Different Play Styles} \label{DP}

The player-specific coefficients $\bbeta_j$ from our regression model (\ref{playermodel}) are what we define as the {\it play style} for each player: how that player's rank and in-match role/game type/map choices affect their performance.  Without any further modeling, we would be estimating a unique $\bbeta_j$ (i.e. unique play style) for each player in our Battlefield 3 data.  

However, we may not have that many games available for each player and we risk over-fitting our match data with so many parameters in our model.   A general solution to over-parameterized models is shrinkage that is encouraged by employing a common prior for the player-specific parameters $\bbeta_j$.   We employ a Dirichlet process (DP) specification for our common prior which encourages clustering of the player-specific parameters $\bbeta_j$, allowing us to discover distinct play styles that are shared by groups of players.  

In this formulation, we assume that the player-specific coefficients $\bbeta_j$ come from a shared but unspecified distribution ${\rm F}$ and then a Dirichlet process prior is specified for ${\rm F}$, 
\begin{eqnarray}
\bbeta_j & \sim & {\rm F} (\cdot) \qquad\qquad j = 1,\ldots,n   \nonumber \\
{\rm F} (\cdot) & \sim & {\rm DP}(\omega, {\rm F}_0) 
\end{eqnarray}
where ${\rm F}_0$ is a prior measure that encapsulates prior beliefs about ${\rm F}$ and $\omega$ is a concentration parameter that specifies the strength of those beliefs in ${\rm F}$.   We provide details about our specification for prior parameters ${\rm F}_0$ and $\omega$ in the next subsection.  

Intuitively, a Dirichlet process prior can be viewed as a less parametric alternative to traditional random effects models, such as $\beta_j \sim ~  \text{Normal} (\mu,\sigma^2)$, where the player-specific $\beta_j$'s would be shrunk towards a single mean $\mu$.   In contrast, a Dirichlet process prior allows for an unspecified number of player-specific parameter means that are shared by subsets of players.   As we will see in the next subsection, the player-specific $\beta_{j}$ for each player $j$ is clustered together with other highly similar player-specific parameters $\beta_i$'s from other players.

Thus, we will be creating a data-driven grouping of players that exhibit similar play styles.  This approach also allows for a subset of players to be left ungrouped from the rest of the population (i.e. players that show a unique play style compared to all other players).  

\subsection{Model Estimation}\label{staticMCMC}

We use Markov chain Monte Carlo to sample from the posterior distribution of the Bayesian semi-parametric model outlined in Sections~\ref{regression}-\ref{DP}.   Specifically, we use a Gibbs sampling algorithm \citep{GemGem84} where each parameter value is iteratively sampled based on the current values of the other parameters.   

The primary step of our Gibbs sampling algorithm is the sampling of a new value for each $\bbeta_{j}$, the play style parameters of player $j$, conditional on the current values of the play styles of other players (which we collectively denote as $\bbeta_{-j}$) and the residual variance $\sigma^2$, 
\begin{eqnarray}
\bbeta_j \,\, \sim \,\,  p(\bbeta_j \, | \, \bbeta_{-j}, \sigma^2, \by) \label{condbetadist}
\end{eqnarray}

With the Dirichlet Process prior outlined in Section~\ref{DP}, the conditional posterior distribution (\ref{condbetadist}) is a mixture of the continuous prior measure ${\rm F}_0$ and $K$ point masses located at each of the $K$ unique values (ie. clusters) in the current set $\bbeta_{-j}$ of sampled play styles for other players in our data \citep{Liu96}.

We now describe this sampling step in more detail using an informed set of initial values for each $\beta_{j}$, though the Gibbs sampling algorithm can also be initialized with random starting values.   We can use the ordinary least squares estimates $\widehat{\bbeta}_j$ from the regression model (\ref{playermodel}) as initial estimates for each player's play style parameters $\bbeta_j$.  We then cluster these player-specific $\widehat{\bbeta}_j$'s using K-means clustering \citep{Mac67} where we initially set the number of these initial clusters using the heuristic $K = \sqrt{n/2}$, where $n$ is the number of players in our dataset \citep{MultivariateAnalysis80}.     

The centers of these initial clusters are also used to estimate the mean $\bmu$ and variance $\Lambda$ of a multivariate normal distribution over these starting clusters.   This multivariate normal distribution is used as the prior measure ${\rm F}_0$ from which we can sample potentially new unique play styles (new cluster centers) during our cluster refinement.    We then replace each player-specific $\widehat{\bbeta}_j$ with its K-means cluster center $\tilde{\bbeta}_k$, so that each set of player-specific parameters $\bbeta_j$ initially take on only one of $K$ unique play style values.   

During each step of our Gibbs sampling algorithm, we revisit the cluster assignment for each player's play style $\bbeta_j$ based on the current set of clusters in the sampled play styles $\bbeta_{-j}$ of all other players.     

Specifically, for the sampling of a particular player's play style parameters $\bbeta_j$, we examine the current partition of the play styles $\bbeta_{-j}$ of all other players except player $j$.  Let's assume this current partition consists of $K$ clusters, where each cluster $k$ has $n_k$ cluster members and a cluster mean of $\tilde{\bbeta}_k$.  For each cluster $k$, we calculate the conditional posterior probability of $\bbeta_j$ belonging to that cluster,  
\begin{eqnarray}
P^j_k \, = \, p(\bbeta_j \, | \, \tilde{\bbeta}_k, \sigma^2, \by) \,\, \propto \,\, n_k \, \cdot \, e^{-\frac{1}{2\sigma^2} \, \sum\limits_{m} \left( y_{jm} - \bX_{jm} \cdot \tilde{\bbeta}_k \right)^2 } \label{condbetaclusterdist}
\end{eqnarray}
where $y_{jm}$ is the log total score of player $j$ in match $m$ and $\bX_{jm}$ are their rank, role, game type, and map covariates for match $m$ as outlined in equation (\ref{playermodel}).  The leading term of $n_k$ in equation (\ref{condbetaclusterdist}) comes from our assumed Dirichlet process prior which gives preferential allocation towards larger clusters \citep{WalJenDic10}.  

So we can see that the conditional posterior probability $P^j_k$ of $\bbeta_j$ belonging to cluster $k$ favors clusters that represent a common play style (i.e. large $n_k$) or that have a play style $\tilde{\bbeta}_k$ that predicts well the observed scores for player $j$, (i.e. small $\sum\limits_{m} ( y_{jm} - \bX_{jm} \cdot \tilde{\bbeta}_k )^2$).  

Through the prior measure ${\rm F}_0$, our Dirichlet process prior formulation also allows for the creation of a new cluster if player $j$ exhibits a particularly unique play style that is not represented well by the existing clusters.   We implement this option in our Gibbs sampling algorithm by also considering a random sample from ${\rm F}_0$ as a candidate value for $\bbeta_j$.  

Specifically, we sample a new unique set of play style parameters $\bbeta_\star$ from the $MVN (\mu,\Lambda)$ distribution that we specified above as our prior measure ${\rm F}_0$.  Then the conditional posterior probability $P^j_\star$ of $\bbeta_j$ taking on this new $\bbeta_\star$ is 
\begin{eqnarray}
P^j_\star \, = \, p(\bbeta_j \, | \, \bbeta_\star, \sigma^2, \by) \,\, \propto \,\, \omega \, \cdot \, e^{-\frac{1}{2\sigma^2} \, \sum\limits_{m} \left( y_{jm} - \bX_{jm} \cdot \bbeta_\star \right)^2 } \label{condbetanewdist}
\end{eqnarray}
We see that the conditional posterior probability $P^j_\star$ of creating a new unique play style for player $j$ is driven by the prior concentration $\omega$ as well as whether that new play style $\bbeta_\star$ predicts well the observed scores for player $j$ (i.e. small $\sum\limits_{m} ( y_{jm} - \bX_{jm} \cdot \bbeta_\star )^2$). 
In our application to Battlefield 3 in Section~\ref{application}, we set the concentration parameter $\omega = 1$.   

Combining all choices from equations (\ref{condbetaclusterdist}) and (\ref{condbetanewdist}) together, player $j$ is then sampled into one of the currently existing $K$ clusters or a potentially new cluster with probabilities proportional to $\bP^j = (P^j_1,...,P^j_K,P^j_*)$.  If player $j$ is sampled into an existing cluster $k$, then their play style $\bbeta_j$ is set equal to the mean of that play style cluster, $\tilde{\bbeta}_k$.  If player $j$ is sampled into the new cluster defined by $\bbeta_\star$, then their play style $\bbeta_j$ is set equal to $\bbeta_\star$.  

During this sweep through all players, any empty clusters are removed and the cluster means $\tilde{\bbeta}_k$ are updated any time that their cluster membership changes.
So we see that each iteration of our Gibbs sampler iteratively refines the current set of common play styles while also allowing the number of play styles to vary.   

We can also sample the residual variance $\sigma^2$ from its conditional posterior distribution as part of our Gibbs sampling procedure.  However, in our current implementation we instead fix $\sigma^2$ equal to a point estimate $\hat{\sigma}^2 = {\rm MSE}$ where ${\rm MSE}$ is the mean squared error from the ordinary least squares estimation of the regression model (\ref{playermodel}).

\section{Application to Battlefield 3} \label{application}

As discussed in Section~\ref{data}, data for the online military game Battlefield 3 was provided by Electronic Arts through the Wharton Customer Analytics Initiative.  The dataset consists of 515,605 player-versus-player match logs taken from 1221 players.   The global regression model (\ref{globalmodel}) has 58 coefficients $\balpha$, which includes the intercept and slope on rank plus 56 coefficients for the different maps, roles and game types that can be chosen by players.   

The player-specific regression model (\ref{playermodel}) adds 58 coefficients $\bbeta_j$ for each player $j$ that represent how their overall performance differs from the average player ($\beta_{j0}$), how they differ from the average player in their change in performance as their rank increases ($\beta_{j1}$), and how their performance differs from average in each role, map, and game type ($\beta_{j2},\ldots,\beta_{58}$). 

Thus, we need to estimate a total of 70876 coefficients (58 global coefficients plus 58 coefficients per player $\times$ 1221 players) as well as the residual variance parameter $\sigma^2$ using the Markov Chain Monte Carlo procedure described in Section~\ref{staticMCMC}.    Our model estimation was implemented in Matlab where parameter initialization via ordinary least squares takes 1.5 minutes and each step of the Gibbs sampler takes approximately 3 minutes on a 64-bit laptop having 8 GB RAM and 2.5 GHz processors.   Our sparse matrix takes 20 MB of disk space and could be optimized to require less space.  

We first provide an overall evaluation of our model fit in Section~\ref{modelevaluation}.   We interpret the clusters of player-specific coefficients from our model to detect common play styles across Battlefield 3 players in Section~\ref{commonplaystyles}, as well as examining {\it hybrid} players that switch between play styles in Section~\ref{hybrid}.  

\subsection{Evaluation of Model Fit}\label{modelevaluation} 

We can evaluate our model fit by examining the out-of-sample (OOS) prediction errors when using a 90-10 hold out scheme where the training set consists of a random 90\% of each player's matches and the test set is the remaining 10\% of each player's matches.  

As a baseline for these evaluations, using a single global average across all players had an OOS root mean square error (RMSE) of 0.97.  The initial regression model with player-specific coefficients (\ref{playermodel}) fit by ordinary least squares has an OOS RMSE of 0.79, which is a 19\% predictive improvement.   The best partition from our MCMC implementation had a similar OOS RMSE of 0.80 but with a substantially reduced number of unique play styles (only 13\% of the unique coefficients compared to the OLS model). 

\subsection{Common Play Styles} \label{commonplaystyles} 

We now shift our attention to interpretation of the parameters of our estimated model, with our first focus being an examination of common play styles exhibited by subsets of players in our Battlefield 3 data.  These common play styles are inferred as the different clusters of player-specific coefficients $\bbeta_j$ in the maximum {\it a posteriori} (MAP) partition found by our Gibbs sampling algorithm.   This MAP partition contains 90 clusters of play styles shared by multiple players.    

\begin{figure}
        \centering
                \includegraphics[scale=0.9]{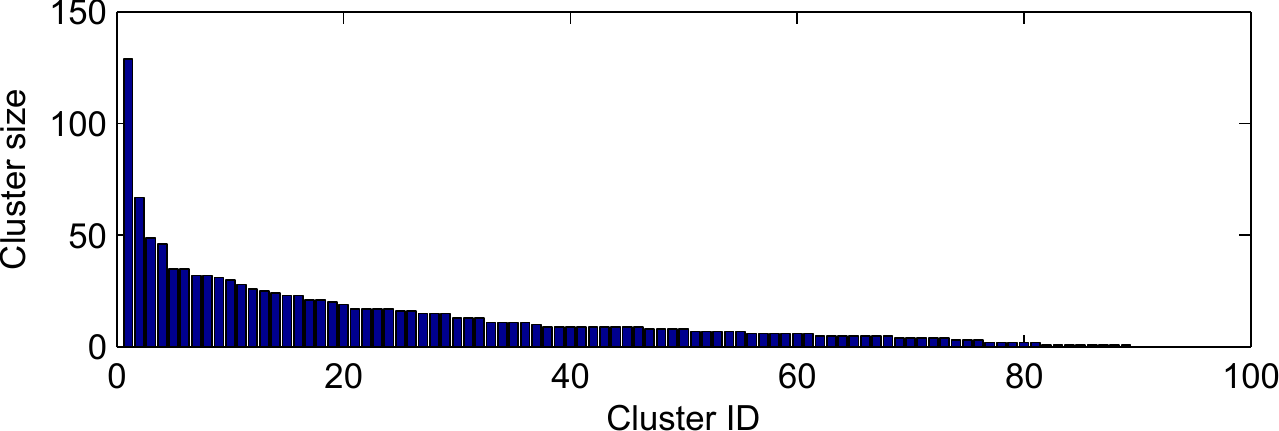}
        \caption{Distribution of cluster sizes for the 90 clusters of multiple players in the MAP partition.} \label{fig:clusters}
\end{figure}

Figure \ref{fig:clusters} gives the distribution of cluster sizes for the 90 clusters of multiple players in the MAP partition.   The largest 30 clusters contain 72\% of the 1221 players in our Battlefield 3 data.

\begin{figure}
\includegraphics[width=\columnwidth]{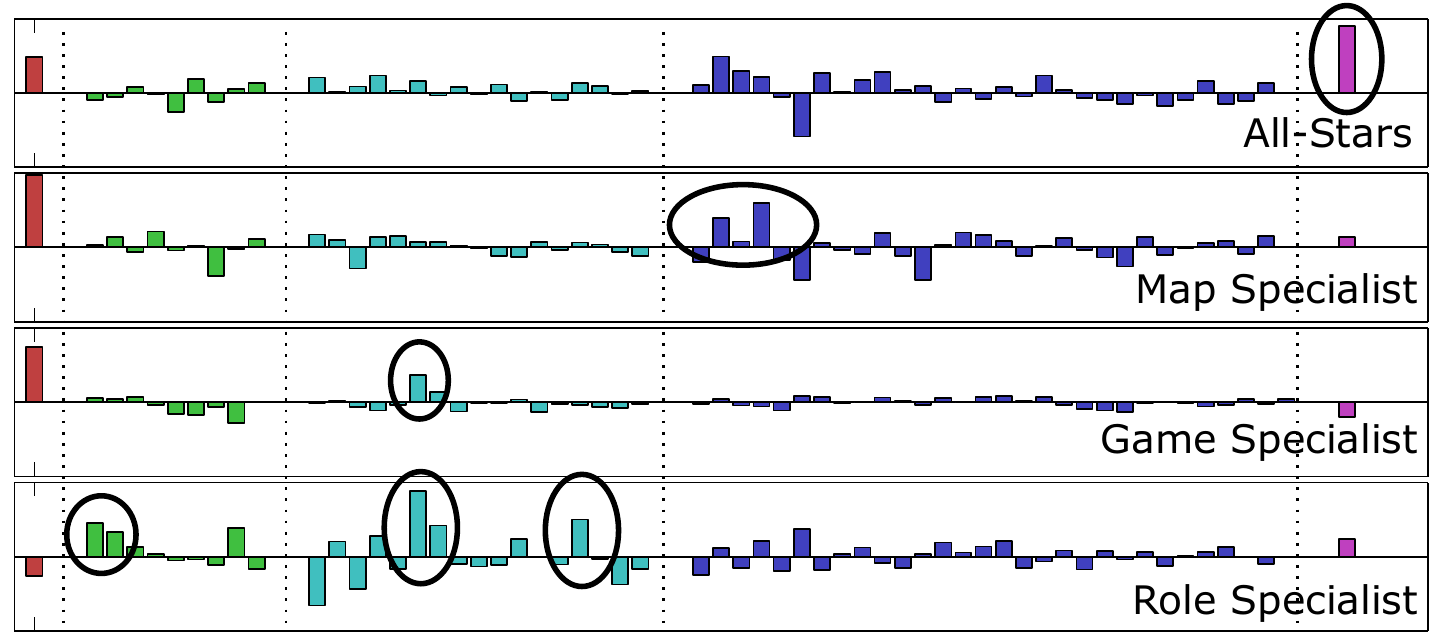}
\caption{Visualization of the four most common play styles (clusters of player-specific coefficients) found by our model. The height of each bar 
represents the magnitudes of the coefficients within that cluster.  The colors indicate the type of feature: player rank (red), role (green), game type (cyan), map (blue), and intercept (purple). \label{ClusterViz}}
\end{figure}

We now examine in more detail some of the most common play styles (i.e. largest clusters) exhibited by players in our Battlefield 3 data.   In Figure~\ref{ClusterViz}, we visualize the four most common {\it play styles} exhibited by players in our data, i.e. the four largest clusters in the MAP partition of the player-specific coefficients across all players. 

Recall that the player-specific coefficients $\bbeta_j$ from our linear regression model (\ref{playermodel}) defines each player's {\it style}: how that player's score is affected by their rank and the roles, games, and maps encountered by that player in their Battlefield 3 matches.   For example, a large player-specific intercept $\beta_{j0}$ indicates that player $j$ performs better than the average player overall, whereas a large positive player-specific coefficient on rank $\beta_{j1}$ indicates that player improves more quickly as they gain ranks than the average player.  

The first cluster in Figure~\ref{ClusterViz} corresponds to players with substantially positive player-specific intercepts, indicating overall performance that is higher than the average Battlefield 3 player.   The other three clusters correspond to players who perform substantially better than the average player within specific combinations of role, game, and map. 

Note that the roles, maps, and game types are inter-related since each map may support only a subset of game types and roles, e.g. if a player excels at a given map, they will often have higher scores for the roles and game types associated with that map.  It is also worth noting that a subset of maps are only available as additional downloadable content and so we have less observed data for this subset of maps.  

\begin{figure*}[ht!]
\includegraphics[width=\textwidth]{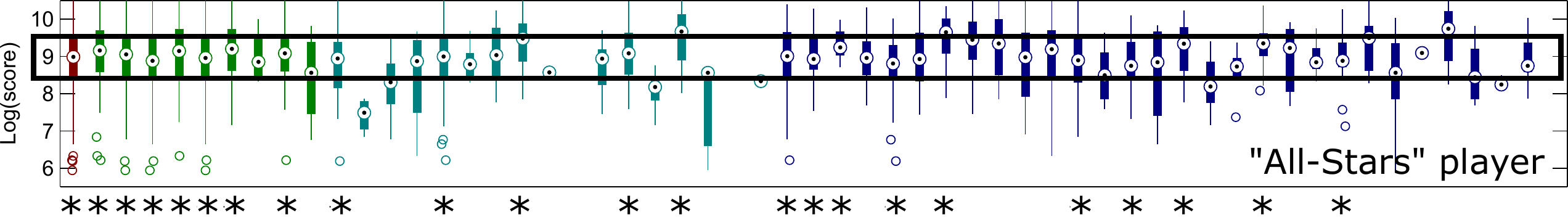}\\
\includegraphics[width=\textwidth]{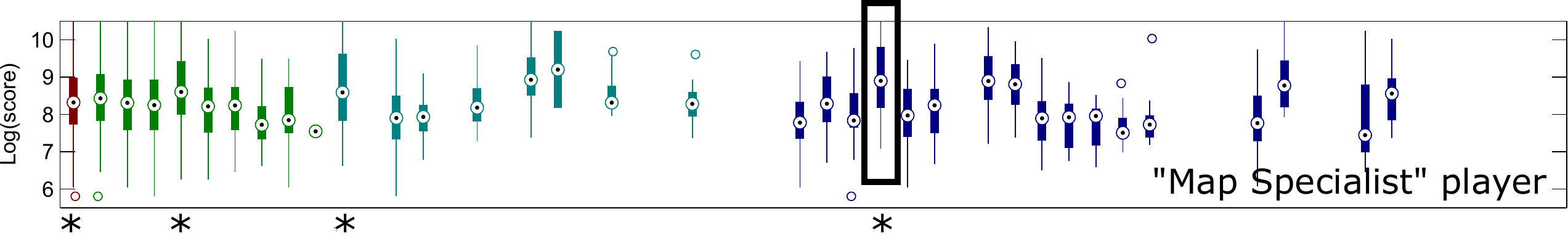}\\
\includegraphics[width=\textwidth]{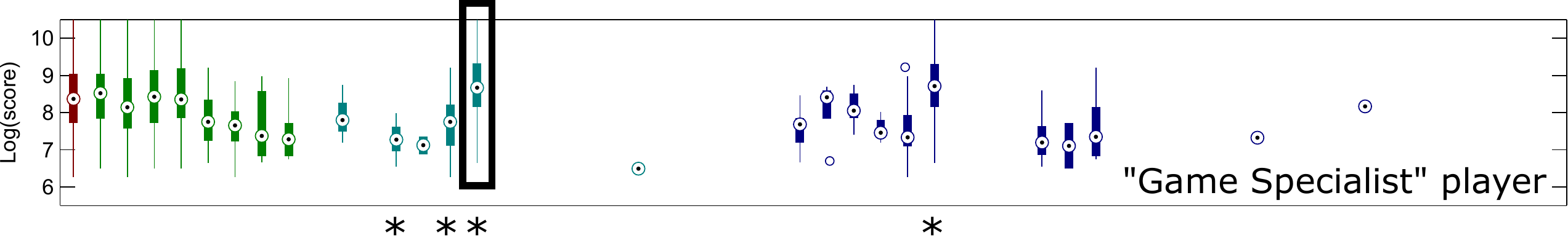}\\
\includegraphics[width=\textwidth]{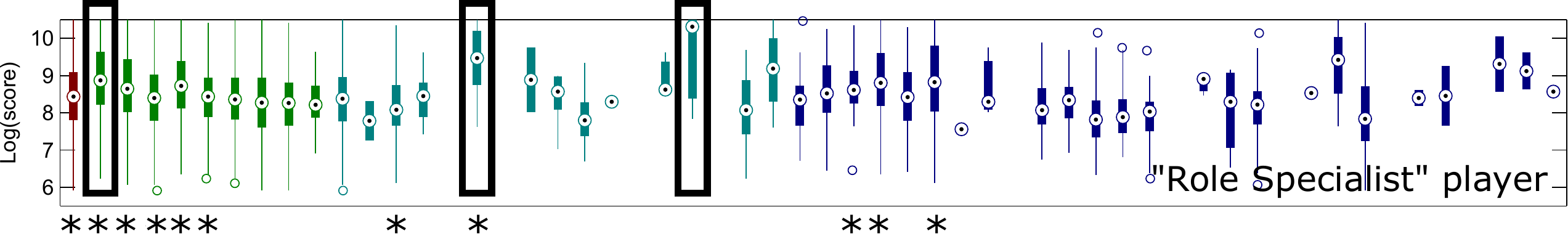}\\
\caption{Box plots of the distribution in performance for 4 players across a variety of different match settings.   From left to right, colors indicate the distribution of log match scores for different subset of matches for that player:  overall/all matches (red), by different role (green), by different game type (cyan), and by different maps (blue).  Blank spaces indicate that the player did not have a match with that particular role, game, or map.  Stars indicate that the player's average log match score is significantly different from the global average (as determined by a t-test).}
 \label{PlayerProfiles}
\end{figure*}

We label the first cluster in Figure~\ref{ClusterViz} as the {\bf All-Stars}: players with a large positive player-specific intercepts $\beta_{j0}$ which indicates that the average game scores in their matches are substantially larger than the average Battlefield 3 player.   The average log score of players in our data was approximately 8 whereas players in this {\bf All-Stars} cluster had average log scores closer to 9.  This group generally plays equally well on all roles, game types, and maps (indicated by relatively small coefficients for roles/games/maps).   This cluster is also the largest found by our model, containing 11\% of the players in our data.  We will examine a representative member of this cluster in Figure~\ref{PlayerProfiles}.

Many clusters found by our model have particular map types as their primary determinants of the variation in their score.  The second cluster in Figure~\ref{ClusterViz} is labelled as the {\bf Map Specialists} since this cluster has large positive coefficient values for two maps, Operation Metro and Grand Bazaar, indicating higher performance for these players than the average Battlefield 3 player in those specific map situations.  This cluster contained about 4\% of players in our data. In Figure~\ref{PlayerProfiles}, we show a representative member of this cluster who consistently performs well in Operation Metro.

Several common play styles found by our model have a particular game type as the main determinant of the variation in their score.  The third cluster in Figure~\ref{ClusterViz} is labelled as the {\bf Game Specialists}, which have a high weight on rank and on two types of team death matches.    This cluster also contained about  4\% of players in our data.  In Figure~\ref{PlayerProfiles}, we show a representative member of this cluster who consistently performs well in one type of death match.

Finally, we label many clusters as {\bf Role Specialists} as they had large coefficients on a particular chosen role along with large coefficients on particular game types or maps that are well suited for that role.   The fourth cluster in Figure~\ref{ClusterViz} is a cluster with the largest positive coefficients on the ``assault" role as well as ``team death match" game type, which incidates better performance than the average Battlefield 3 player in those game and role situations.   This cluster contained about 3\% of the players in our data.  Figure~\ref{PlayerProfiles} shows a representative member of this cluster, who performs significantly better than average in one team death match type as well as assault.

In Figure~\ref{PlayerProfiles}, we give examples of four player profiles, each one being a representative member of the four clusters shown in Figure~\ref{ClusterViz}.   
The {\bf All-Stars} player in Figure~\ref{PlayerProfiles} tends to perform above average across almost every role, game and map (recall that the average log score of players in our data was approximately 8 whereas this player had average log scores closer to 9).   The {\bf Map Specialist} player displays their best performance in certain maps, in this case the Operation Metro map, as well as certain game types associated with that map.   The {\bf Game Specialist} player consistently performs well in death match, which is one particular game type.  The {\bf Role Specialist} player in Figure~\ref{PlayerProfiles} performs especially well in the assault role as well as the corresponding team death match game type.  

We note that all four players shown in Figure~\ref{PlayerProfiles} had player-specific coefficients that tended to remain in the same clusters throughout the iterations of the Gibbs sampler.   However, some players in our Battlefield 3 data had player-specific coefficients that frequently moved between different clusters, which would indicate a change in their {\it play style} over time.   We examine players with these ``hybrid" play styles in the next Section~\ref{hybrid}.  

\subsection{Hybrid Play Styles: Players that frequently change clusters} \label{hybrid}

As outlined in Section~\ref{staticMCMC}, our model estimation allows each player to ``move" between different {\it play style} by sampling their player-specific coefficients $\bbeta_j$ into different clusters of unique values across iterations of the Gibbs sampler.   The transitions (or lack thereof) that a player makes between clusters gives an indication of how well that player fits any particular common {\it play style}. 

\begin{figure}
\includegraphics[scale=0.75]{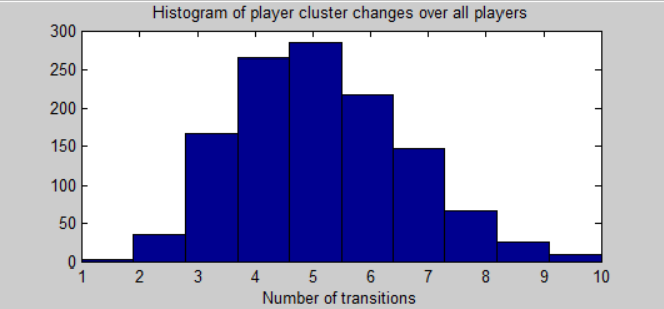}
\caption{The distribution of the number of different player-specific coefficient clusters visited by players in our Battlefield 3 data}\label{transition}
\end{figure}

In Figure~\ref{transition}, we show the distribution of the number of different play styles exhibited by the players in our Battlefield 3 data.  

We want to contrast players whose cluster memberships do not change very often across MCMC iterations (`stable' players) versus players whose cluster memberships change frequently across MCMC iterations (`hybrid' players).   In this analysis, we define players who remain in the same cluster for over 50\% of the Gibbs sampler iterations as `stable'.  With this criterion, 28\% of the players in our Battlefield 3 data are stable players that consistently display the same common play style. All four representive players that are displayed in Figure~\ref{PlayerProfiles} are stable members of their respective play style clusters.  

We define hybrid players as those that transitioned between the same two clusters at least 4 times over all iteration of the Gibbs sampler.  Under this criterion, 7\% of players in our Battlefield 3 data are defined as hybrid players.  In some cases, hybrid players belong to similar clusters, i.e. two different clusters that both have large coefficient values on the same feature.    

A small group of players (3\%) transitioned betweeen 9 different clusters of play style coefficients.   Most of these players played fewer than 30 games at beginner ranks and so our model may not have enough data to estimate a more precise and stable play style for these players.   
 
\section{Discussion} \label{discussion}

The complexity of game play in online multiplayer games such as Battlefield 3 has spurred a tremendous amount of interest in modeling different play styles and strategies towards victory.   In particular, identifying subgroups of participants that use similar play styles would aid game developers in developing specialized tutorials for new participants as well as improving the construction of complementary teams in their online matching queues.  

We contribute to this endeavor by developing a hierarchical Bayesian regression approach for Battlefield 3 players that models total game score as a function of player rank as well as the roles, game type, and map taken on by that player in each of their matches.   Player-specific intercepts account for differences in overall player ability while player-specific coefficients for the other covariates can be interpreted as the strenths or weaknesses of each player under specific role, game and map choices.  

We use a Dirichlet process prior that enables the clustering of players that have similar player-specific coefficients, which allows us to discover common {\it play styles} amongst our sample of Battlefield 3 players.  This flexible semi-parametric Bayesian clustering approach does not require the number of distinct play styles to be known {\it a priori} and also allows for a subset of players to be left ungrouped with their own unique play styles.  These characteristics are important for this application since prior information is lacking about the landscape of Battlefield 3 play styles.  

In terms of overall predictive performance, the ordinary least squares version of our regression model has 19\% better predictive accuracy compared to a baseline global average across all players whereas using the Dirichlet process prior leads to similar predictive accuracy but with a substantial reduction in the number of parameters (only 13\% of the unique coefficients compared to the OLS model).

We examine several of the most common play styles among Battlefield 3 players and find a set of ``all-star" players that exhibit high overall performance, as well as groupings of players that perform particularly well in specific game types (``game specialists") or specific maps (``map specialists"), as well as groupings based on performance in specific roles (e.g. healing or assault).   We are also able to differentiate betweeen players that are stable members of a particular play style from ``hybrid" players that exhibit multiple play styles across their matches.  

Future work will investigate whether these discovered groupings can help players improve their own performance or help matchmaking algorithms to form better teams of players with complimentary skill sets.   We also plan to explore using additional game features as well as experimentiing with different outcome variables beyond total game score.

\section{Acknowledgments}

We wish to thank Electronic Arts and the Wharton Customer Analytics Initiative (WCAI) for their feedback and support.

\bibliographystyle{imsart-nameyear}
\bibliography{references}

\end{document}